# Behavioural Correlation for Detecting P2P Bots

Yousof Al-Hammadi
The Intelligent Modelling & Analysis Research (IMA),
School of Computer Science and Information Technology,
Jubilee Campus, The University of Nottingham,
Nottingham, NG7 1BB, UK
E-mail: yxa@cs.nott.ac.uk.

Uwe Aickelin
The Intelligent Modelling & Analysis Research (IMA),
School of Computer Science and Information Technology,
Jubilee Campus, The University of Nottingham,
Nottingham, NG7 1BB, UK
E-mail: uxa@cs.nott.ac.uk.

## Abstract

In the past few years, IRC bots, malicious programs which are remotely controlled by the attacker through IRC servers, have become a major threat to the Internet and users. These bots can be used in different malicious ways such as issuing distributed denial of services attacks to shutdown other networks and services, keystrokes logging, spamming, traffic sniffing cause serious disruption on networks and users. New bots use peer to peer (P2P) protocols start to appear as the upcoming threat to Internet security due to the fact that P2P bots do not have a centralized point to shutdown or traceback, thus making the detection of P2P bots is a real challenge. In response to these threats, we present an algorithm to detect an individual P2P bot running on a system by correlating its activities. Our evaluation shows that correlating different activities generated by P2P bots within a specified time period can detect these kind of bots.

## 1. Introduction

Internet and networks come under frequent attack from a diverse set of malicious programs and activity such as viruses and worms [3]. While the detection of such worms and viruses is improving a new threat has emerged in the form of the botnet. Botnets are distributed networks of infected machines, controlled remotely by a 'botmaster'. A single bot, a term derived from robot, is a piece of malicious code that is installed on a user host and transforms host into a zombie machine.

Current bots use Internet Relay Chat (IRC) command and control (C&C) structure to communicate with their herders. IRC is a chat based protocol consisting of various 'channels' to which a user of the IRC network can connect. Upon infection of a host, the bot connects to the IRC server and joins the specified channel waiting for the attacker's commands. The bot is programmed to respond to various commands generated by the attacker through the C&C structure. Although IRC structures represent an efficient way of controlling botnets, one can prevent the bots from communicating with their herders by shutting down the central point. In order to avoid this problem, botnet herders started to deviate from using a centralised point to a another way of controlling their bots by using the decentralised structures as a mean to maintain their botnets. As a result, a new approach of botnet structures starts to appear taking advantage of existing Peer-to-Peer (P2P) protocols. The attackers start to use P2P networks in order to control their botnets. By using this approach, the bots can contact other bots without having a centralised point for their command and control (C&C) structure.

In P2P, each node acts as client-server which provides bandwidth, storage and computational power. Using this approach, bots are able to communicate with other bots by downloading files or commands from other bots' machines and performing different activities. In comparison to IRC structures, everyone can join a P2P network, thus, the more peers acting as bots, the more powerful the botmaster can be. In addition, it will be hard to detect and shut down the botnet as security people would need to isolate each machine [2].

We present an algorithm to detect P2P bots on the infected machine by correlating bots' behavioural attributes. A Peacomm (Storm P2P bot) is used as a case study. The concept of correlation attributes within specified time-window increases the level of malicious behaviour confidentiality as depending on one process attribute may generate large number of false alarms. The correlation is used to combine the attributes of programs, thus, enhance the detection mechanism. The algorithm does not need a pre-defined bots' signatures in order to detect this kind of bots.

This paper is structured as follows: Section two discusses the related work in this field. We present the Peacomm P2P bot as a case study in Section three. We discuss our methodology by explaining the conducted experiments, how we have collected our data and the correlation algorithm in section four. Our results and analysis are presented in section five and we summarize and conclude in section six.

## 2. Related Work

Different researches have been conducted to analyse and detect peer-to-peer bots. For example, Schoof and Kon-ing [10] analyse different peer-to-peer bots such as Sinit and Nugache to examine the behaviour of these bots. In their analysis, they note that some peer-to-peer bots communicate on a fixed port. They argue that by monitoring traffic on that port, one could detect these bots. They also discover that some of these bots generate a large number of destination unreachable error messages (DU) and connection reset error messages while trying to connect to other peers. In addition, some bot's communications are encrypted which make the traffic analysis a difficult task and resulting in high false alarms.

Dittrich and Dietrich [1] explain some of the features and challenges when dealing with the Nugache P2P botnet. Stover et al. [12] conclude that there is no static IDS that will detect Nugache traffic. They also mention that the Nugache bot can be detected through various signatures of the infection.

Holz et al. [8] present a method to analyse and mitigate P2P botnet. They develop ways to mitigate Storm worm and introduce an active measurement technique to enumerate the number of infected hosts. Their way is based on either reverse engineered the bot binary to identify the function which generates the key that is used for searching for other infected machines and bots commands or use honeypot and infect it with the bot that generate a new key each time it is rebooted and thus enumerate all the keys. The problem with the first method is that the process of reverse engineering is needed all the time the attacker change the key generation functions. The other method takes long time to enumerate all the keys.

Other research analyse different peer-to-peer bots such as Storm bot (Peacomm) where large number of emails are spammed to many accounts holding an executable attachment [2]. An in-depth analysis of Peacomm is provided by Stewart from SecureWorks [11]. Nunnery and Kang [7] try to locate the zombie nodes activities in peer-to-peer network by their retrieval of hashes and the control of a large group of network computers. They claimed that if the client within the controlled network searches for hash used by malware, it must be a zombie node. This process can also leads to locate the IP address of the botmaster by monitoring any publish activity on the supervised network.

A detailed description of Peacomm is presented by Porras et al. [9]. They also investigate how to detect the Storm bot by using a BotHunter [5] which tracks the two-way communication flows between internal and external entities to find the infected host. Stover [12] suggests that the Storm bot can be detected by configuring IDS to find the configuration file used by the bot. They also state that differentiating between the Storm bot and legitimate P2P communications is a difficult task.

## 3. Case Study: Peacomm (Storm Bot)

Peacomm is one of the few known bots which implements a full peer-to-peer (P2P) networks generated in 2007 [7][4]. Peacomm uses the Overnet P2P protocol to communicate with other bots. The overnet protocol is an example of a decentralised peer network which implements the Kademlia algorithm [7]. The Kademlia algorithm uses distributed hash tables (DHT) for routing in which there is no hierarchy in the topology [4][7].

In Kademlia, to communicate with other peers, each node is assigned a unique 128-bit ID as a global identifier the first time the it starts [8]. The node shares its information with other peers which are close to it in the keyspace. The term "closeness" is based on calculating the distance between two peers' identifications (IDs) using an XOR operation [6][8]. To publish information, each node uses its hash table as a data structure which maps the keys with the values. The key is used as an identification to retrieve information based on the closest distance to other peers while the value is a triple of node's ID, IP address and UDP port number. The Peacomm bot uses this method to search for specific keys to communicate with other infected machines on the network and find the commands that should be executed [8].

After an infection, the bot stores a configuration file which contains encoded information about other peers to communicate with [8]. Each time it makes a successful connection, it expands its information about the botnet by saving the hash (node information) into local memory. The bot then uses the stored hard coded keys to search and download a secondary injection URL from the network. The secondary injection URL is encrypted using the RSA algorithm and points to a secondary injection executable. After that, the bot decrypts the encrypted URL using its stored hard coded keys. Finally, the bot downloads and execute the secondary injection from a web-server [7][4].

## 4. Methodology

P2P bots are difficult to detect because they do not have a central point to communicate with their masters which makes the tracing back process difficult. In addition, new bots can have different signatures and using signature-based detection will generate false alarms specially if bots traffic are encrypted. For these reasons, we present an algorithm to detect P2P bots by investigating the effect of correlating their behavioural's attributes. The aim of this investigation is to show that correlating different attributes can enhance the detection of P2P bots.

### 4.1. Bot Scenarios

For the purpose of our experimentation the Peacomm bot is used as previously explained. As a communication vessel,

IceChat, an IRC client, is used for normal conversation. In addition, the Firefox web browser is used for browsing, checking email and other normal activities as a normal application. Two scenarios are used inactive (PmE1) and active (PmE2) as follows:

- inactive (PmE1): In this session, the binary of Peacomm bot is executed and run on a monitored host. Other normal applications are also running during this session but there are no activities from the user such as browsing or chatting.
- active (PmE2): In this session, the Peacomm bot is executed and run on a monitored host. In contrast to (PmE1), the user uses Firefox for browsing and checking emails as normal activities and uses Icechat for having conversation with other users.

### 4.2. Data Collection

We assume that the bot is already installed on the victim host, through an accidental 'trojan horse' style infection mechanism or opening a malicious attachment which contains the bot. In this case, we use an extrusion detection to limit the bot activities whilst on a host machine. The communication of Peacomm bot is described in more details in [8][4][12].

We have developed an interception program (APITrace) to record the required behavioural attributes and to intercept and capture specified function calls executed by the monitored processes. These data are then processed, normalised and streamed to our correlation algorithm. In terms of the function calls intercepted, different types of function calls are used as an input to the algorithm. These function calls include Communication functions (e.g. send, recv), File access functions (e.g ReadFile, WriteFile), Registry access functions (e.g. RegOpenKey, RegQueryValue) and Keyboard status functions (e.g. GetKeyboardState, GetAsynKeyStat).

### 4.3. Signals

Three signal categories are used to define the state of the system namely $S_1$, $S_2$ and $S_3$. These signals are collected using a function call interception program - APITrace. Raw data from the monitored host are transformed into log files which are then normalised in the range of 0 - 100.

In terms of the signal category semantics, $S_1$ is a strong signal evidence for bad behaviour on a system. This signal is derived from the rate of change of three fields of netstat. These fields are destination unreachable (DU), failed connection attempts (FCA) and reset connections (RST). The choice of these fields is based on the preliminary observation of P2P bots. The netstat is a command line tool used to display network statistics. The value of $S_1$ is obtained from the following formula:

$$S_1 = (DU_t - DU_{t-1}) + (FCA_t - FCA_{t-1}) + (RST_t - RST_{t-1})$$

The normalisation of data is based on a logarithmic scale. This is because we needed to cover a data of large range of values being produced by flooding attack and these values are changing rapidly. If the value of the $S_1$ exceeds 100, the value is capped to a maximum value, in our case 100. Otherwise, the value of $S_1$ is calculated as shown in Table 1.

Table 1. Values of $S_1$ for P2P experiments

| DU  | RST | FCA | PAMP                  |
|-----|-----|-----|-----------------------|
| 0   | 0   | 0   | 0                     |
| 0   | 0   | !=0 | 100*log10(RST)        |
| 0   | !=0 | 0   | 100*log10(FCT)        |
| !=0 | 0   | 0   | 100*log10(DU)         |
| 0   | !=0 | !=0 | 100*log10(RST+FCA)    |
| !=0 | 0   | !=0 | 100*log10(DU+FCA)     |
| !=0 | !=0 | 0   | 100*log10(DU+RST)     |
| !=0 | !=0 | !=0 | 100*log10(DU+RST+FCA) |

$S_2$ is derived from the rate of change of number of packets send per second (pkts/sec). This value is also obtained from network statistics command line tool (netstat). Based on preliminary experiments, we classify the values of $S_2$ according to the following:

$$X = \begin{cases} 0 - 10 & min\ danger \\ 11 - 100 & min\text{-}mid\ danger \\ 101 - 1000 & mid\ danger \\ 1001 - 10000 & mid\text{-}max\ danger \\ > 10000 & max\ danger \end{cases}$$

The higher the value of $S_2$, the more threat we have. We also use a logarithmic scale when using $S_2$ and is derived according to the following formula:

$$S_2 = 25 * \log 10(X) \ldots 1 \leq X \leq 10,000$$

If the rate of change of the number of packets sent per second exceeds 10,000, the value of $S_2$ is capped to 100. In addition, if the rate of change is zero, the value of $S_2$ is mapped to zero.

Finally, $S_3$ is derived from the time difference between two outgoing consecutive communication functions such as [(send,send),(sendto,sendto),(socket,socket)]. This is needed as the bot either sends information to the botmaster using send function call or issues SYN or UDP flooding attacks using sendto or socket. In normal situations, we expect to have a large time difference between two consecutive functions. In addition, we expect to have a short period of this action in comparison to a SYN attack or a UDP attack. This is because the behaviour of the user when responding to other parties is different from bots when responding to their botmaster commands. Often, the normal users do not involve in generating large amount of traffic similar to flooding attack when they chat or when they browse the Internet.

Therefore, we set $n_{s3}$ as the maximum time differences between calling two consecutive communication functions. If the time difference is higher than $n_{s3}$, the time is classified as normal and it is mapped to a 100. If the time difference falls below $n_{s3}$, the time difference is calculated from the following formula:

$$S_3 = 62.41965 * log10(Y)$$

The value of 62.41965 is calculated from preliminary experiments in which we observe that the average safe value between calling the same consecutive communication function calls is around 40 and any value above this value can be considered to be as normal. Therefore, if Y = 40, the value of $S_3$ is capped to 100 which represent the normal situation. The closer the value of $S_3$ to 0, the more malicious activity we have.

The need of correlation between signals and function calls is required to define which processes are active when the signal values are modified. The more active the process, the more function calls it generates. Once the function calls are intercepted by APITrace, they are stored in different log file and assigned the value of the process ID to which the function calls belong and the time at which they were invoked. After a certain period of time, both signal and function calls logs are combined and sorted based on time. The combined log files are parsed and the logged information is sent to the correlation algorithm for processing and analysis.

4.4. The Correlation Algorithm

We have implemented a correlation algorithm to find the correlation between the three signals $S_1$, $S_2$ and $S_3$. The correlation method is based on the two criteria. The first criteria is to analyse the function calls log file based on the frequency of API function calls generated by each process (i.e. calculate the number of function calls invoked per process). The second criteria is to analyse the signals log file by setting a sensitivity value (SV) for each signal ($S_1$, $S_2$ and $S_3$).

The algorithm is described in Algorithm 1 works as follows. We set a sensitivity value (SV) and check if the values of $S_1$, $S_2$ and $S_3$ exceeds the specified a SV. If the signal value exceeds the specified SV, we assign a value of one to its records, otherwise, we assign a value of zero. Then, we examine if signals' records have the same values, we assign a value of one which represents a correlation between the signals ($S_1$, $S_2$ and $S_3$) at that period of time. We repeat this process for all the signals in the signals log file.

Then, we calculate the anomaly factor and the correlation factor from the following equations:

$$AnomalyFactor(AF) = \sum_{i=1}^{n} \frac{(X_{S1i} + X_{S2i} + X_{S3i})}{3n} \quad (1)$$

```
input : S= (S₁, S₂, S₃)
Initialise SV;
for i = 1 to n do
    if S1ᵢ > SV then
        | X_{S1i} = 1;
    else
        | X_{S1i} = 0;
    end
    if S2ᵢ > SV then
        | X_{S2i} = 1;
    else
        | X_{S2i} = 0;
    end
    if S3ᵢ > SV then
        | X_{S3i} = 1;
    else
        | X_{S3i} = 0;
    end
    if X_{S1i} = 1 and X_{S2i} = 1 and X_{S3i} = 1 then
        | Corrᵢ = 1;
    end
end
```
**Algorithm 1**: A correlation algorithm

$$CorrelationFactor(CorrF) = \sum_{i=1}^{n} \frac{Corr_i}{n} \quad (2)$$

where n is the time in seconds and X is the signal record which represents a logic value (zero or one) if the signal value exceeds a predefined sensitivity value (SV). The correlation factor represents how signals are related to each other and its range from zero to one. For example, if $S_1$ and $S_2$ have high values than sensitivity value (SV) and $S_3$ has a low value than (100-SV), (note that signal values are normalised from zero to 100, thus we change the SV from zero to 100.), this will generates a high correlation between these signals at that time. The final step is to calculate the anomaly correlation value (ACV) from the following equation:

$$ACV = AF * \exp(CorrF) \quad (3)$$

The use of exponential form to the correlation factor in this formula represents the confidentiality level of how signals are related to each other. For example, if the correlation factor is zero, this means that the signals in the log files are not correlated and the ACV will only depends on the anomaly factor (AF). If the correlation factor is higher than zero, the ACV will depend on both the anomaly factor and the correlation factor. Thus, the more correlation we have between signals, the higher the ACV will be. The maximum value of ACV is 2.7182 which is the value of exp(O) as the value of anomaly factor ranges from zero to one as well.

## 5. Results and Analysis

The results of applying this technique are shown in Table 2. In this table, the frequency of API function calls for each process for the conducted experiments and the anomaly correlation values (ACV) when applying different SV are presented (SV=10,20,30,40,50). As shown from this table, we note that changing SV value generates different anomaly correlation values (ACV). If we increase the sensitivity of the system by increasing the SV, this will lead to the reduction of ACV as shown in Table 2.

To detect malicious activity, a threshold value is needed. Setting a threshold value T = 50 detects malicious activities on system for experiments PmE1 and PmE2 when SV ranges from 10 to 30 but cannot detect malicious activity on the system when we increase the value of SV to 40 or more. Therefore, setting a threshold level in the range 10 to 30 will generate zero false positive alarms and 100% true positive alarms for the two conducted experiments.

Another important question is to know which processes are malicious and which processes are normal. Using the frequency of API function calls for each process as an indicator, it will determine which process is normal and which process is malicious in this case study. For example, based on the frequency of API function calls, the Peacomm is a malicious process in experiments PmE1 and PmE2.

Table 2. The results of using the correlation algorithm when (1) applying different sensitivity values (SV) to calculate the Anomaly Correlation Value (ACV) and (2) considering the frequency of API function calls per process for P2P bots.

| Experm. | Process | Freq. | ACV (>SV) | | | | |
|---|---|---|---|---|---|---|---|
| | | | 10 | 20 | 30 | 40 | 50 |
| PmE1 | Peacomm Firefox Icechat | 617349 9902 71 | 0.68 | 0.60 | 0.52 | 0.45 | 0.34 |
| PmE2 | Peacomm Firefox Icechat | 628838 4449 9464 | 0.69 | 0.63 | 0.56 | 0.44 | 0.33 |

## 6. Conclusion

P2P bots are difficult to detect as there is no central point of communications. In addition, analysing network traffic looking for signatures can be tedious task because bots signatures can be dynamic and encrypted. In this work, we have developed a correlation algorithm to detect bots on the system by correlating their behavioural activities. Our results show that correlating different activities can enhance the detection mechanisms and reduce the false alarms. One disadvantage of this algorithm is that the value of threshold to detect malicious processes is undefined and further experiments are needed to set a proper threshold for detecting malicious activity in the system. In addition, different types of P2P bots should be examine to verify the accuracy of the correlation algorithm.

## Acknowledgment

The authors would like to thank Khalifa University of Science, Technology And Research (KUSTAR) - UAE, for providing financial support for this work.